\documentclass[sigconf]{acmart}

\usepackage{algorithm}
\usepackage{algorithmic}
\usepackage{amsmath}
\usepackage{amsfonts}
\usepackage{multirow}
\usepackage{multicol}
\usepackage{graphics}
\usepackage{caption}
\usepackage{subcaption}
\usepackage{mathtools}

\def\ourMethod{Diverse Deep Feature Ensemble Learning}
\def\ourMethodAbrv{D$^2$FEL}

\AtBeginDocument{%
  }

\copyrightyear{2024}
\acmYear{2024}
\setcopyright{rightsretained}
\acmConference[ICMIP 2024]{2024 9th International Conference on Multimedia
and Image Processing (ICMIP)}{April 20--22, 2024}{Osaka, Japan}
\acmBooktitle{2024 9th International Conference on Multimedia and Image
Processing (ICMIP) (ICMIP 2024), April 20--22, 2024, Osaka, Japan}
\acmDOI{10.1145/3665026.3665036}
\acmISBN{979-8-4007-1616-4/24/04}

\begin{document}
\begin{sloppypar}
\title[\ourMethod{} for ODG-ReID]{\ourMethod{} for Omni-Domain Generalized Person Re-identification}

\author{Eugene P.W. Ang}
\authornote{Corresponding author.}
\orcid{0000-0001-7507-6569}
\affiliation{%
  \institution{Rapid-Rich Object Search (ROSE) Lab, Nanyang Technological University}
  \streetaddress{50 Nanyang Ave, S2-B4b-13}
  \city{Singapore}
  \country{Singapore}
  \postcode{639798}
}
\email{phuaywee001@e.ntu.edu.sg}

\author{Shan Lin}
\orcid{0000-0002-3254-1923}
\affiliation{%
  \institution{Rapid-Rich Object Search (ROSE) Lab, Nanyang Technological University}
  \streetaddress{50 Nanyang Ave, S2-B4b-13}
  \city{Singapore}
  \country{Singapore}
  \postcode{639798}
}
\email{shan.lin@ntu.edu.sg}

\author{Alex C. Kot}
\orcid{0000-0001-6262-8125}
\affiliation{%
  \institution{Rapid-Rich Object Search (ROSE) Lab, Nanyang Technological University}
  \streetaddress{50 Nanyang Ave, S2-B4b-13}
  \city{Singapore}
  \country{Singapore}
  \postcode{639798}
}






\renewcommand{\shortauthors}{Ang et al.}

\begin{abstract}
Person Re-identification (Person ReID) has progressed to a level where single-domain supervised Person ReID performance has saturated. However, such methods experience a significant drop in performance when trained and tested across different datasets, motivating the development of domain generalization techniques. However, our research reveals that domain generalization methods significantly underperform single-domain supervised methods on single dataset benchmarks. An ideal Person ReID method should be effective regardless of the number of domains involved, and when test domain data is available for training it should perform as well as state-of-the-art (SOTA) fully supervised methods. This is a paradigm that we call \textit{Omni-Domain Generalization} Person ReID (ODG-ReID). We propose a way to achieve ODG-ReID by creating deep feature diversity with self-ensembles. Our method, \ourMethod{} (\ourMethodAbrv{}), deploys unique instance normalization patterns that generate multiple diverse views and recombines these views into a compact encoding. To the best of our knowledge, our work is one of few to consider omni-domain generalization in Person ReID, and we advance the study of applying feature ensembles in Person ReID. \ourMethodAbrv{} significantly improves and matches the SOTA performance for major domain generalization and single-domain supervised benchmarks.
\end{abstract}


\maketitle

\section{Introduction}
Person Re-identification (ReID) is the task of matching images of a person captured across multiple non-overlapping cameras. In single-domain ReID, methods are trained and tested on the same multi-camera system (domain) with mutually exclusive person identities between training and testing sets. A key disadvantage of conventional single-domain ReID methods is that they generally perform worse when tested on other domains (e.g., the mAP drops from 76.3 to 16.1 as shown in Table~\ref{tab:test}), motivating the development of cross-domain ReID methods that assume the training and testing sets come from different domains. This is a step toward practical application as it approximates real-world constraints such as having limited or no access to data in the target domain. However, cross-domain methods can have overly specialized requirements such as having access to multiple source domains for training. Table~\ref{tab:test} reveals that state-of-the-art (SOTA) cross-domain methods do well when tested on unseen domains but under-perform simple baselines when testing on the original source domains. For example, when a recent SOTA method ACL~\citep{ACL-DGReID} is trained on a mix of datasets that include DukeMTMC-reID, it can only achieve a mAP of 72.9 when tested on the DukeMTMC-reID test set, which is lower than the simple Bag-of-Tricks (BoT) baseline performance~\citep{StrongBaseline}. Our goal for this work is to investigate Person ReID methods that can achieve strong performance regardless of domain setting. The ideal ReID methods should perform well in both single domain setting and cross-domain setting, while still performing well in the original training domain. We refer to such methods as Omni-Domain Generalized ReID (ODG-ReID) methods.

\begin{table}[h]
\center
\caption{The performance of single-domain methods degrades when testing on other datasets, while cross-domain generalization methods cannot maintain the same performance on the original training dataset. All results are in mAP and the best result for each benchmark is bold. Dataset abbreviations: M is Market-1501~\citep{Market-1501}, D is DukeMTMC-reID~\citep{DukeMTMC-reID}, C is CUHK03~\citep{CUHK03} and MS is MSMT17$\_$V2~\citep{MSMT17}.}
\resizebox{\columnwidth}{!}{
\begin{tabular}{c|c|cc|cc}
\hline
\multirow{2}{*}{Setting} & \multirow{2}{*}{Methods} & \multicolumn{2}{c|}{Single Domain Train} & \multicolumn{2}{c}{Mult-Domain Train} \\ \cline{3-6} 
& & \multicolumn{1}{c|}{\textbf{D} → \textbf{D}}  & M → \textbf{D} & \multicolumn{1}{c|}{C+\textbf{D}+M → \textbf{D}} & C+M+MS → \textbf{D} \\ \hline
Normal & Baseline (BoT)~\cite{StrongBaseline} & \multicolumn{1}{c|}{\textbf{76.3}} &  16.1   & \multicolumn{1}{c|}{\textbf{76.3}}         &     38.6      \\ \hline
\multirow{3}{*}{DG} & DualNorm~\cite{DualNorm}       & \multicolumn{1}{c|}{67.7} &   23.9  & \multicolumn{1}{c|}{76.3}         &  40.5         \\
                         & META~\cite{META-DGReID}           & \multicolumn{1}{c|}{46.7} &  18.3   & \multicolumn{1}{c|}{60.6}       &  42.7         \\
                         & ACL~\cite{ACL-DGReID}            & \multicolumn{1}{c|}{49.9} &  22.2   & \multicolumn{1}{c|}{72.9}     & \textbf{53.4}          \\ 
                         & SIL~\cite{StyleInterleaved} & \multicolumn{1}{c|}{61.5} & \textbf{25.1} & \multicolumn{1}{c|}{65.6} &  47.0 \\ \hline
\end{tabular}%
}


\label{tab:test}
\end{table}

\begin{figure}[t]
\centering
\includegraphics[width=0.9\columnwidth]{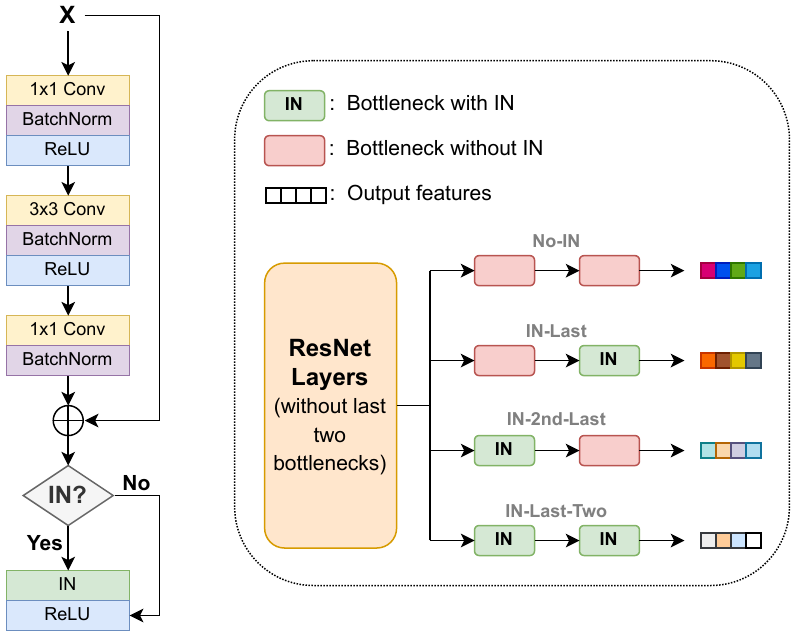}

\caption{Left: Structure of a bottleneck layer in a ResNet~\citep{ResNet}, where IN can be selectively applied at the end. Right: all possible IN-combinations applied to the final two bottlenecks of the network.}
\label{fig-IN-recipes}
\end{figure}

Our method re-investigates the use of Instance Normalization (IN), a technique popularly used in Domain Generalization Person ReID (DG-ReID) methods~\citep{DualNorm}. IN normalizes the style attributes of images so that models learn domain-invariant features and ignore domain-specific style cues that lead to overfitting. First proposed for DG-ReID in DualNorm~\cite{DualNorm}, IN layers were added to the early blocks of the network but not to the later layers because the former seemed to perform better. While this design decision is superior in some benchmarks, we uncover a more complicated story: unconventional applications of IN can also achieve superior performance in other benchmarks. Table~\ref{tab:in-position} illustrates how IN is applied in the bottlenecks of a ResNet~\citep{ResNet} and enumerates all possible IN-combinations on the final two bottlenecks. Trying these four combinations on two example benchmarks, the table in Figure~\ref{fig-IN-recipes} shows that adding IN in the later bottlenecks can be beneficial. In this table, different IN patterns excel in different benchmarks regardless of domain setting. Ultimately, no single IN pattern is superior in all cases. Based on these findings, a successful method has to consider different IN patterns \textit{all at the same time} in order to do well for unseen target benchmarks. Our method, the \ourMethod{} (\ourMethodAbrv{}), uses self-ensembling to achieve this simultaneous diversity: each subhead applies a unique IN pattern and they collectively view the input through the lens of all possible IN-combinations. 

\begin{table}[h]
\center
\caption{Various IN-patterns are applied to the final bottleneck layers of a backbone ResNet-50 and each pattern is evaluated on same-domain and cross-domain benchmarks. No single pattern (column) outperforms the rest consistently in all benchmarks (row). Results are in Rank-1 and the best of each benchmark is bold. Columns correspond to the branch labels in Figure~\ref{fig-IN-recipes}.}
\resizebox{\columnwidth}{!}{
\begin{tabular}{c|cccc} \hline
                 & \textbf{No-IN} & \textbf{IN-Last} & \textbf{IN-2nd-Last} & \textbf{IN-Last-Two} \\ \hline
\textbf{M $\rightarrow$ M}  & 92.5            & \textbf{94.3}             & 93.1             & 94.2             \\
\textbf{C $\rightarrow$ C}  & \textbf{90.6}            & 89.0             & 88.1             & 90.4             \\
\textbf{MS $\rightarrow$ MS}  & 79.0            & 79.3             & 79.4             & \textbf{80.5}             \\
\textbf{D $\rightarrow$ MS}  & \textbf{41.1}            & 35.4             & 38.9             & 36.0             \\
\textbf{M $\rightarrow$ MS}  & 30.6            & 30.3             & \textbf{31.4}             & 31.0             \\
\textbf{C $\rightarrow$ MS} & 33.4            & 35.3             & 33.0             & \textbf{37.6}     \\ \hline       
\end{tabular}
}
\label{tab:in-position}
\end{table}

Self-ensembles are a resource-efficient way to perform ensemble learning. As Figure~\ref{fig-ourmethod-big} illustrates, the subheads in self-ensembles are copies of the original network head sharing the early layers of a feature extraction backbone. Such a structure reduces memory use and enables training of the ensemble all in one go. Initially proposed for image classification tasks~\citep{one-knowledge-distil-on-the-fly-ensemble} and used to perform knowledge distillation in ReID~\citep{RESL-self-ensemble-ReID}, we found no works that studied how to directly apply self-ensembles for inference in ReID. The usual way to apply ensemble techniques in image classification is to average the predicted class logits from different subheads as the ensemble's own prediction. A naive extension to ReID would be to average the output \textit{features} of all subheads instead, using the averaged feature as a search key. Experimentally, this yields poor results and the reason for this is that it makes sense to average class logits, but not features. In the former case, each logit component corresponds to the confidence score for a particular person's identity. Taking the average of these logits makes sense because there is a perfect correspondence between logits produced by different models. In the latter case, however, features from different subheads are not necessarily aligned along their coordinate axes and averaging them degrades performance, as Figure~\ref{fig-compare-logits-features-avg} (left) illustrates and the experiment on the right confirms. We found that it was better to concatenate all features together to preserve their information. To the best of our knowledge, our work is the first to propose the application of self-ensembles in Person ReID inference.

As the number of subheads increase, concatenating their features together can be impractical due to the large features produced. We propose an effective way to reduce their dimensions using the Johnson-Lindenstrauss (JL) lemma~\cite{JL_Lemma}. The JL-lemma relates the dimension reduction of a set of data points to the conservation of pairwise distances between the same points in the reduced space. By performing a random projection from a high dimension $D$ down to a lower dimension $d$, we only sacrifice a small amount of performance even when the reduction is very wide, e.g. $D=16384, d=2048$. No training cost is incurred, making this a cost-effective method. One disadvantage of random projections, however, is that as $d$ gets very small the quality of the reduced features drops sharply.

If minimal performance loss is desired, we found that traditional dimension reduction models designed to minimize information loss such as Principal Component Analysis (PCA) work very well. Such principled techniques work well regardless of the training set provided (assuming a sufficient sample size), generalizing well to unseen domains with minimal loss in performance. Furthermore, PCA-reduced features maintain strong performance even down to very small dimensions, making it ideal for applications that need lean search features. Training a PCA model only incurs a one-time effort after \ourMethodAbrv{} is trained and is relatively fast.

Finally, our experiments reveal that deep learning dimension reduction techniques such as auto-encoders perform well in same-domain settings but perform poorly in cross-domain settings. This suggests that deep-learning approaches to dimension reduction have a tendency to overfit to the training set by exploiting domain specific information during the encoding process, but ignore important domain generalizable details. We propose traditional reduction techniques since they are mathematically formulated to preserve information during reduction. Table~\ref{tab:ablation-autoenc-vs-randproj-pca} performs a comparison of dimension reduction techniques across various benchmarks.

To summarize, our contributions are as follows:
\begin{itemize}
    \item We propose \ourMethodAbrv{}, a framework that extracts rich and diverse views of data with a self-ensemble architecture. To the best of our knowledge, we are the first to apply feature ensembles directly for Person ReID inference and search.
    \item We pioneer the novel adoption of the principled \textit{Random Projection} reduction technique into Person ReID, enabling practical use of feature ensembles in ReID.
    \item Our method \ourMethodAbrv{} outperforms the state-of-the-art (SOTA) in several multi-source domain generalization Person ReID benchmarks while still matching or surpassing the SOTA performance for single domain Person ReID benchmarks, demonstrating Omni-Domain Generalization.
\end{itemize}

\section{Related Work}
\textbf{Single-Domain Person Re-identification}
Single-domain Person ReID benchmarks, which train and test models within a single multi-camera system (domain), are among the earliest efforts of the Person ReID community. Such modern methods are mostly based on deep convolutional neural networks. Early deep learning based work usually formulated Person ReID as a verification problem and proposed Siamese architectures \cite{CUHK03, Huang2019Multi-PseudoRe-Identification, Lin2017End-to-EndRe-Identification}. In recent years, verification-driven triplet architectures such as \cite{TriNet, Quadruplet} overtook Siamese structures due to their robustness. Another popular approach uses the classification loss: Zheng et al.~\cite{IDE} first proposed the ID-Discriminative Embedding (IDE) to finetune ImageNet~\cite{ImageNet} pretrained models to classify person IDs. More recent approaches \cite{MGN,StrongBaseline,AGW-pami21reidsurvey} combine both classification loss and verification loss. 

\noindent\textbf{Domain-Generalized Person ReID}: A task that has been gaining traction in the field of Person ReID, Domain-Generalized Person ReID (DG-ReID) aims to learn a model that can generalize well to unseen target domains without involving any target domain data for adaptation. IBN-Net~\citep{IBN-Net} first introduced the instance normalization (IN) layer to normalize the style and content variations within the image batch during the training. This was later applied to DG-ReID by DualNorm~\citep{DualNorm}. MMFA-AAE~\citep{MMFA-AAE} used a domain adversarial learning approach to remove domain-specific features. Later DG methods such as DIMN~\citep{DIMN}, QAConv~\citep{QAConv}, M$^3$L~\citep{M3L} used hyper-networks or meta-learning frameworks coupled with memory bank strategies. DEX~\citep{DEX} proposed a deep feature augmentation loss that implicitly transforms intermediate deep features over expected Gaussian noise perturbation. ACL~\citep{ACL-DGReID} propose a module that captures then fuses domain invariant and domain specific features, which is plugged into different layers of a generic feature extractor as a replacement for selected conv blocks. Style Interleaved Learning (SIL)~\cite{StyleInterleaved} is a recent method that employs an additional forward pass during training to interleave feature styles. Finally, methods like RaMoE~\citep{RaMoE-Dai2021} and META~\citep{META-DGReID} deploy a mixture of experts to specialize to each domain, with META being a more lightweight variant because of extensive parameter sharing. Our method \ourMethodAbrv{} is most similar in concept to mixture of experts because we deploy a diverse mix of specialized network subheads.

\noindent\textbf{Self-Ensembles}
Self-ensembles were first proposed by ONE~\cite{one-knowledge-distil-on-the-fly-ensemble} for image classification, which extended a base network with auxiliary branches that each generate logits to be used as soft labels for knowledge distillation. RESL~\cite{RESL-self-ensemble-ReID} applied a similar architecture for domain adaptation in Person ReID, also using it for knowledge distillation. In both methods, the auxiliary branches are discarded after knowledge distillation in order to reduce computation. Our method differs from them because we use self-ensembles not for knowledge distillation, but as a way to extract rich and diverse features from the input. Our work is the first to explore how self-ensemble features can be directly used for inference and search in Person ReID instead of discarding the auxiliary branches.

\noindent\textbf{Dimension Reduction}
The JL-Lemma tells us that when a set of high-dimensional points are randomly projected to a lower dimension, their pairwise distances are preserved up to a bounded error that depends mainly on the size of the target dimension~\citep{JL_Lemma, JL_Lemma_Gupta1999AnEP}. The solid theoretical foundations of the JL-Lemma gives strong guarantees of the feasilbility of random projections as a real-time dimension reduction tool. In this study, we demonstrate the potential of random projections as a practical component in a Person ReID framework. Principal component analysis (PCA)~\cite{pca-pearson, pca-Hotelling1933AnalysisOA} linearly projects datasets into smaller subspaces while retaining the directions of greatest variance.


\section{Methodology}
We are given a dataset of person images with identity labels $D=\{x_i,y_i\}_1^n$. A model $f$ maps images to feature encodings - for this work, we fix $f$ to be a ResNet-50, which is made up of bottleneck components as illustrated in Figure~\ref{fig-IN-recipes}. We refer to these bottlenecks using indices counting backwards from the output of the network. For example, bottleneck-1 is the last layer and bottleneck-2 comes before bottleneck-1. 
\subsection{Instance Normalization (IN) Patterns}
A key feature of \ourMethodAbrv{} is the ability to apply varying patterns of IN to the subhead bottlenecks. Given an input batch $x \in \mathbb{R}^{B \times C \times H \times W}$ with $b$ indexing batch size $B$, $c$ indexing number of channels $C$, spatial dimensions $H,W$, and a small value $\epsilon$ for numerical stability, an IN operation is specified as:
\begin{align}
    \text{IN}(x_{bchw}) &= \frac{x_{bchw} - \mu_{bc}}{\sqrt{\sigma^2_{bc} + \epsilon}}, \\
    \mu_{bc} &= \frac{1}{HW} \sum_{i=1}^{W} \sum_{j=1}^{H} x_{bcij}, \\  
    \sigma^2_{bc} &= \frac{1}{HW} \sum_{i=1}^{W} \sum_{j=1}^{H} (x_{bcij} - \mu_{bc})^2
    \label{eq:instance-norm}
\end{align}
In each bottleneck layer, we have a choice to apply IN or not. Our experiments indicate that using a diverse set of IN combinations generates a richer set of features that consistently perform well across extensive Person ReID benchmarks. We introduce the concept of Fully-Combinatorial-IN for a depth $\delta \in \mathbb{Z}^{+}$ as the simultaneous application of all $2^\delta$ IN-combinations on bottlenecks-$(\delta,\delta\!-\!1,...,1)$. Figure~\ref{fig-IN-recipes} illustrates Fully-Combinatorial-IN for $\delta=2$. The table in Fig~\ref{fig-IN-recipes} compares the effect of changing IN-patterns in the last two bottlenecks over several benchmarks, showing that it is impossible to predetermine patterns that work for all target domains. Thus, a Fully-Combinatorial-IN strategy spanning all patterns is the best solution to this uncertainty. One of the ways we can implement Fully-Combinatorial-IN is to train multiple models, each using a different IN pattern. Our method \ourMethodAbrv{} takes a leaner approach and achieves this with self-ensembles.
\subsection{Our Method: \ourMethodAbrv{}}
\begin{figure}[t]
\centering
\includegraphics[width=1.0\columnwidth]{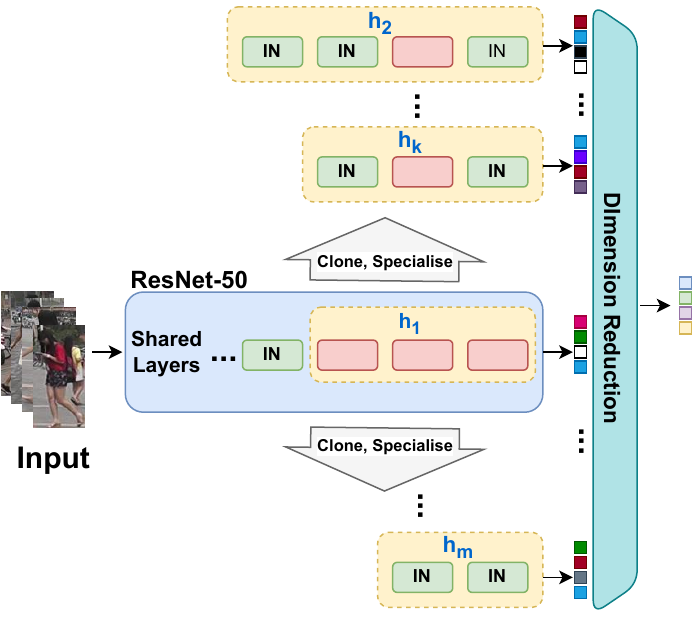}
\caption{A summary of our \ourMethodAbrv{} framework, which uses partial copies of itself to form an ensemble. Each copy is specialized with a unique pattern of instance normalization.}
\vspace{-1.5em}
\label{fig-ourmethod-big}
\end{figure}

Figure~\ref{fig-ourmethod-big} illustrates our method, \ourMethodAbrv{}. We fix the main backbone $\mathbf{f}$ and clone $m-1$ subheads $\{\mathbf{h}_k\}_2^m$ from the final bottlenecks of $\mathbf{f}$ for a total of $m$ heads. All subheads share the earlier layers of $\mathbf{f}$. For each subhead, we can specify the start point $\{\mathbf{f}_k\}_2^m$ to duplicate from and the IN-pattern. This design gives the flexibility to express just a subset of all possible IN patterns for more lightweight requirements. We define $\mathbf{f} \coloneqq \mathbf{h}_1 \circ \mathbf{f}_1$. Given an input image $x$, the \ourMethodAbrv{} function $\mathbf{\omega}$ produces a collection of features, one from each of the $m$ heads:
\begin{equation}
    \mathbf{\omega}(x) = \{\mathbf{h}_k(\mathbf{f}_k(x))\}_1^m
\end{equation}
\begin{figure*}[t]
\centering
\includegraphics[width=2.0\columnwidth]{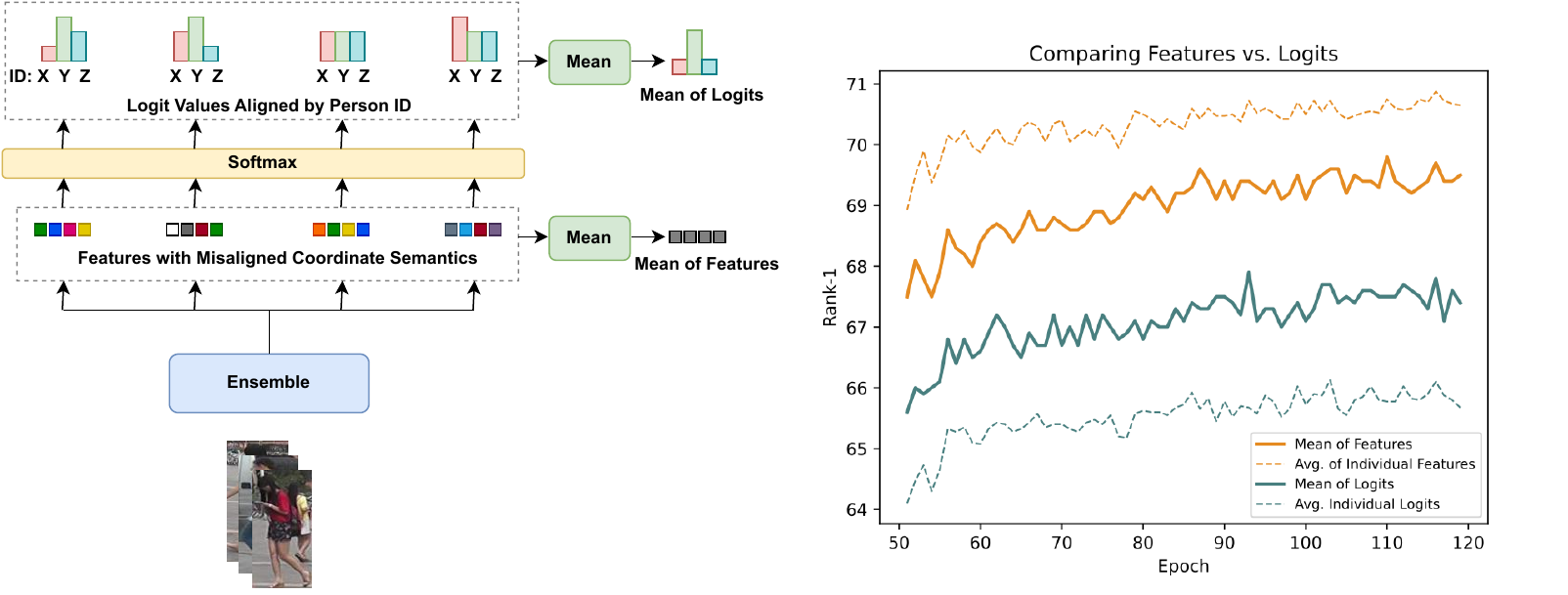} 
\caption{We compare the effects of averaging the coordinate values of class logits vs. regular output features. Even though logits do not perform as well as features for ReID, a pattern is clear: averaging class logits boost performance because they are semantically aligned by design, but averaging features reduces performance because their coordinates do not necessarily align.}
\vspace{-1em}
\label{fig-compare-logits-features-avg}
\end{figure*}
For classification tasks, these subheads are usually linearly projected into class logits and averaged for knowledge distillation or testing. In Person ReID, we want to directly use the features for retrieval instead. Figure~\ref{fig-compare-logits-features-avg} demonstrates, however, that taking average of features in this naive manner is detrimental: the performance of the averaged feature is no better than individual subhead features. Instead, if we were to use the \textit{logits} as retrieval vectors, the performance of the average of all logits is far better than that of an individual logit. This is because logit components are aligned with each other, as each component refers to the same person identity across all subheads, while feature components do not necessarily follow such alignment. Therefore, instead of taking a mean of the features we concatenate them together into larger features:
\begin{equation}
    \Omega(x) = \big\lVert_1^m h_k(f_k(x))
\end{equation}
While $\Omega(x)$ can be directly used for the Person ReID search, as the number of subheads increases it can grow very large, motivating the need for good dimension reduction techniques to efficiently reduce $\Omega(x)$ to a practical size.
\subsection{Dimension Reduction}
A set of $m$ omni-normalized features from \ourMethodAbrv{} each with 2048-dimensional features would have a combined dimension of $D = 2048m$ that needs to be reduced to be practical. We found that deep learning solutions for dimension reduction, such as deep auto-encoders, performed well in single-domain Person ReID but not as well across domains, indicating an overfit to the training domain. Instead, we found that older methods in the literature, such as principal component analysis and random projections, were more robust to ever-changing domain settings. Furthermore, to the best of our knowledge, we are the first to adapt random projection techniques, traditionally applied in the field of database search, for Person ReID. We present our study of these techniques in the following sections.

\subsubsection{Deep Auto-Encoders}
\begin{figure}[ht]
\centering
\includegraphics[width=0.9\columnwidth]{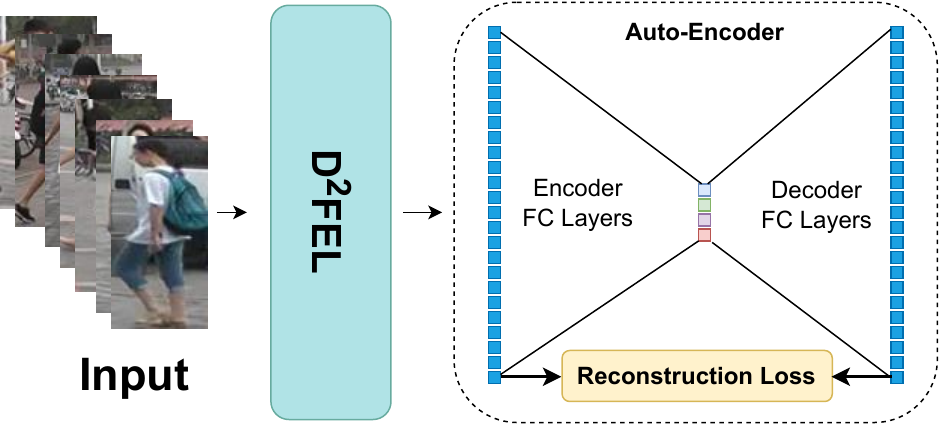} 
\caption{Design of our deep auto-encoder experiments.}
\vspace{-1em}
\label{fig-deep-autoencoder}
\end{figure}

Figure~\ref{fig-deep-autoencoder} illustrates the basic design of the deep auto-encoder experiments we conducted for dimension reduction. We train the auto-encoder on the feature ensemble generated by \ourMethodAbrv{}, using either $L_1$ or $L_2$ reconstruction loss. We use the compressed feature (coloured squares in the middle) for evaluation. Table~\ref{tab:ablation-autoenc-vs-randproj-pca} shows that deep learning dimension reduction methods overfit to the training set, performing well in single-domain Person ReID but performing poorly across domains.
\subsubsection{Principal Component Analysis (PCA)}PCA constructs an orthogonal basis from the eigenvectors of $\{\Omega(x)\}_1^n$ that maximally captures the information encoded in the features~\citep{pca-pearson,pca-Hotelling1933AnalysisOA}. By selecting the eigenvectors corresponding to the $d$ largest eigenvalues, PCA maximally captures the diversity of the data in $X$ using only $d$ dimensions. While the training time of PCA grows with $D$, it is still only a fraction of the time taken to train deep learning models. Furthermore, this is a one-time cost - once trained, it can be deployed at the end of \ourMethodAbrv{}, as illustrated in Figure~\ref{fig-ourmethod-big}. Compared with random projections, Figure~\ref{fig-randproj-pca-chart} shows that PCA can preserve performance down to much smaller dimensions. Thus, PCA is an ideal solution if maximal performance or small but effective search features are desired.

\subsubsection{Random Projections}
Given a set of high dimensional features $\{\Omega(x)\}_1^N \in \mathbb{R}^D$, the Johnson-Lindenstrauss (JL) lemma states the conditions necessary for a random projection from dimension $D$ to $d \ll D$ to preserve the pairwise distances between points in $\mathbb{R}^D$ and their projected counterparts in $\mathbb{R}^d$. If these distances are preserved, any evaluation performed on the projected points in $\mathbb{R}^d$ would be very close to evaluating the original high-dimensional points. A random projection matrix $U \in \mathbb{R}^{D \times d}$ can be constructed independently from data: $U_{ij} \sim \mathcal{N}(0,1), \forall i,j$. We can define the projection $\Pi \in \mathbb{R}^D \times \mathbb{R}^d$ as: $\Pi(\Omega(x)) = U^{\intercal}\Omega(x)$. If $d \geq 24 \ln{N}(3\epsilon^2 - 2\epsilon^3)^{-1}$, then we can in randomized polynomial time find a projection $\Pi$ such that the pairwise distances are preserved to within a tolerance $0 \leq \epsilon \leq 1$ for all $u,v \in \{\Omega(x)\}_1^N$~\cite{JL_Lemma_Gupta1999AnEP}:
\begin{equation}
    (1-\epsilon)\lVert u - v \lVert^2 \, \leq \, \lVert \Pi(u) - \Pi(v) \lVert^2 \, \leq \, (1+\epsilon) \lVert u - v \lVert^2
\end{equation}
For example, if the gallery we are searching has $N=10000$ points and we wish to reduce feature dimensions down to $d=2048$, then the JL-lemma guarantees the existence of a projection that achieves the error bound $\epsilon = 0.2$. For random projections, $\epsilon$ is independent of the original dimension $D$, which is a very useful property since the time taken to fit other dimension reduction models like PCA grows cubically with $D$, but random projections do not require training and achieve error rates of $\epsilon$ as long as we fix the target dimension $d$. While speed, convenience and performance are major plus points of random projections, there are trade-offs to consider. Firstly, there will still be a slight drop in performance. Secondly, the target dimension needs to be reasonably large otherwise performance drops severely (Figure~\ref{fig-randproj-pca-chart}), making random projections ill-suited for applications that require very small search features.

\subsection{Loss Function}
For each ensemble output feature $\omega_k(x)$ with label $y$, we apply three losses. First, we apply a cross-entropy loss over the person identity labels: $L_{CE} = -\log \frac{\exp{\mathbf{w}_y \cdot \omega_k(x)}}{\sum_{j=1}^{C}\exp{\mathbf{w}_j \cdot \omega_k(x)}}$, where $\mathbf{w}$ are classifier weights indexable by label $y$ and $C$ is the total number of classes. We also apply triplet loss: $L_{Tri} = [d_{ap} - d_{an}]_+$, where $d$ is a distance metric and $a,p,n$ are anchor, positive and negative samples in a batch. Finally, we apply center loss~\cite{Wen2016ARecognition}: $L_{Cen} = \lVert \omega_k(x) - \mathbf{c}_y \rVert_2$, where $c_y$ is a learnable centroid of all features belonging to class $y$. Using weights $0 \leq \alpha_1,\alpha_2,\alpha_3$, the overall loss is the weighted sum of these three losses over all $m$ subheads:
\begin{equation}
    L_{D^2FEL} = \sum_{k=1}^{m} \left[ \alpha_1 L_{CE} + \alpha_2 L_{Tri} + \alpha_3 L_{Cen} \right]
\end{equation}
One of the advantages of \ourMethodAbrv{} is the simplicity in loss design. By leveraging on powerful and proven loss techniques, \ourMethodAbrv{} avoids over-specialization and achieves all-rounded strong performance across single and cross domain Person ReID benchmarks. 

\section{Experiments}
\label{sec:experiment-settings}

\begin{table*}[ht]
\center
\caption{Results on the modern DG-ReID benchmarks. For the methods with the dagger symbol $\dag$, we evaluated the official open source implementation on this benchmark. \textbf{Bold} numbers are the best, while \underline{underlined} numbers are second.}
\resizebox{1.95\columnwidth}{!}{
\begin{tabular}{l|c|cc|cc|cc|cc|cc}
\hline
\multirow{2}{*}{Method} &
  \multirow{2}{*}{Training Data} &
  \multicolumn{2}{c|}{C+D+MS→M} &
  \multicolumn{2}{c|}{C+M+MS→D} &
  \multicolumn{2}{c|}{C+D+M→MS} &
  \multicolumn{2}{c|}{D+M+MS→C} &
  \multicolumn{2}{c}{\textbf{Average}}\\ \cline{3-12} 
          &            & Rank-1 & mAP  & Rank-1 & mAP  & Rank-1 & mAP  & Rank-1 & mAP & Rank-1 & mAP  \\ \hline
DualNorm~\citep{DualNorm} $\dag$  & Train Only & 78.9   & 52.3 & 68.5   & 51.7 & 37.9   & 15.4 & 28.0   & 27.6 & 53.3 & 36.8 \\
QAConv~\citep{QAConv}    & Train Only & 67.7   & 35.6 & 66.1   & 47.1 & 24.3   & 7.5  & 23.5   & 21.0 &  45.4  & 27.8 \\
OSNet-IBN~\citep{OSNet} $\dag$ & Train Only & 73.4   & 45.1 & 61.5   & 42.3 & 35.7   & 13.7 & 20.9   & 20.9 & 47.9 & 30.5 \\
OSNet-AIN~\citep{OSNet} $\dag$ & Train Only & 74.2   & 47.4 & 62.7   & 44.5 & 37.9   & 14.8 & 22.4   & 22.4 & 49.3 & 32.3 \\
M$^3$L~\citep{M3L}    & Train Only & 75.9   & 50.2 & 69.2   & 51.1 & 36.9   & 14.7 & 33.1   & 32.1 & 53.8  & 37.0 \\
DEX~\citep{DEX}       & Train Only & 81.5   & 55.2 & 73.7   & 55.0 & 43.5   & 18.7 & 36.7   & 33.8 & 58.9  & 40.7 \\
META~\citep{META-DGReID} $\dag$  & Train Only & 66.7   & 44.6 & 61.8   & 42.7 & 32.4   & 13.1 & 21.3   & 21.6 & 45.6 & 30.5 \\
ACL~\citep{ACL-DGReID} $\dag$      & Train Only & \underline{86.1}   & \textbf{63.1} & 71.7   & 53.4 & 47.2   & 19.4 & 35.5 & 34.6 & 59.3  & 41.5 \\ 
SIL~\citep{StyleInterleaved} $\dag$ & Train Only  & 79.2   & 52.8 & 68.3   & 47.0 &   36.9   & 13.8 & 29.5   & 28.9 & 53.5 & 35.6 \\
PAT~\citep{ni2023part} & Train Only   & 75.2   & 51.7 & 71.8   & 56.5 & 45.6   & \textbf{21.6} & 31.1   & 31.5 & 55.9 & 40.3 \\
\hline \hline
CBN~\citep{CBN-ReID} & Train+Test & 74.7   & 47.3 & 70.0   & 50.1 & 37.0   & 15.4 & 25.2   & 25.7 & 51.7  & 34.6 \\
SNR~\citep{SNR}       & Train+Test & 75.2   & 48.5 & 66.7   & 48.3 & 35.1   & 13.8 & 29.1   & 29.0 & 51.5  & 34.9 \\
MECL~\citep{MECL-ReID}      & Train+Test & 80.0   & 56.5 & 70.0   & 53.4 & 32.7   & 13.3 & 32.1   & 31.5 & 53.7  & 38.7 \\
RaMoE~\citep{RaMoE-Dai2021}     & Train+Test & 82.0   & 56.5 & 73.6   & 56.9 & 34.1   & 13.5 & 36.6   & 35.5 & 56.6  & 40.6 \\
MixNorm~\citep{MixNorm}   & Train+Test & 78.9   & 51.4 & 70.8   & 49.9 & 47.2   & 19.4 & 29.6   & 29.0 & 56.6  & 37.4 \\
MetaBIN~\citep{choi2021metabin}   & Train+Test & 83.2   & 61.2 & 71.3   & 54.9 & 40.8   & 17.0 & \textbf{38.1} & 37.5 & 58.4  & 42.7 \\
\hline
\ourMethodAbrv{}-3-RP (Ours) & Train Only & 85.2   & 61.1 & \underline{75.0}   & \underline{57.6} & \underline{48.0}   & \underline{20.8} & 37.0 & \underline{38.5} & \underline{61.3} & \underline{44.5} \\
\ourMethodAbrv{}-3-PCA (Ours) & Train Only & \textbf{86.5}   & \underline{62.3} & \textbf{76.1}   & \textbf{58.2} & \textbf{48.1}   & \underline{20.8} & \underline{37.6}   & \textbf{39.6} & \textbf{62.1} & \textbf{45.2} \\ \hline
\end{tabular}%
}
\label{tab:dg-reid-modern}
\end{table*}

\begin{table*}[ht]
\caption{Results on the Single Domain ReID Benchmarks. For the methods with the dagger symbol $\dag$, we evaluated the official open source implementation on this benchmark. \textbf{Bold} numbers are the best, while \underline{underlined} numbers are second.}
\center
\resizebox{2.0\columnwidth}{!}{
\begin{tabular}{cc|ccc|cccc|ccc} \hline
\multicolumn{2}{c|}{\multirow{2}{*}{\textbf{Method}}}                                       & 
\multicolumn{3}{c|}{\textbf{Same Domain}} & \multicolumn{4}{c|}{\textbf{Domain Generalization}} & \multicolumn{3}{c}{\textbf{Omni-Domain Generalization}}    \\
\multicolumn{2}{c|}{}                                                                       & 
BoT $\dag$ & MGN $\dag$ & AGW & DualNorm $\dag$ & SIL $\dag$ & META $\dag$ & ACL $\dag$ & OSNet-AIN $\dag$ & \ourMethodAbrv{}-RP & \ourMethodAbrv{}-PCA \\ \hline
\multirow{2}{*}{\textbf{Market-1501}} 
& mAP & 80.6 & 85.3 & \textbf{88.2} & 76.4 & 71.9 & 47.2 & 73.1 & 78.5 & 87.2 & \underline{87.6} \\
& R1  & 92.3 & 94.5 & \textbf{95.3} & 91.9 & 88.9 & 75.1 & 88.0 & 92.5 & \underline{95.0} & \underline{95.0} \\ \hline
\multirow{2}{*}{\textbf{DukeMTMC-reID}}                                               
& mAP & 63.0 & 76.3 & \textbf{79.6} & 67.7 & 61.5 & 46.7 & 49.9 & 69.9 & 77.7 & \underline{78.6} \\
& R1  & 78.9 & 86.9 & \underline{89.0} & 83.8 & 79.9 & 73.1 & 71.5 & 84.7 & \underline{89.0} & \textbf{89.5} \\ \hline
\multirow{2}{*}{\textbf{CUHK03}} 
& mAP & 88.5 & 87.6 & 62.0 & 67.2 & 78.1 & 56.0 & 79.8 & 70.0 & \underline{88.6} & \textbf{89.9} \\
& R1  & 90.5 & 91.1 & 63.6 & 68.9 & 82.5 & 60.5 & 83.2 & 74.9 & \underline{91.7} & \textbf{93.0} \\ \hline
\multirow{2}{*}{\textbf{MSMT17 (V2)}} 
& mAP & 49.2 & 50.5 & 49.3 & 43.9 & 37.2 & 37.1 & 27.7 & 42.7 & \underline{57.2} & \textbf{61.0} \\
& R1  & 74.7 & 74.8 & 68.3 & 74.4 & 66.2 & 69.1 & 57.0 & 71.5 & \underline{83.4} & \textbf{83.8} \\ \hline
\end{tabular}
}
\label{tab:reid-single-domain}
\end{table*}

\noindent\textbf{Domain Generalization Person ReID Benchmarks}
\noindent There are a few evaluation benchmarks for DG-ReID methods. Works such as M$^3$L~\citep{M3L} and DEX~\citep{DEX} adopt a leave-one-out evaluation, where three out of four datasets (Market-1501, DukeMTMC-reID, CUHK03 and MSMT17$\_$V2, which we abbreviate to M, D, C and MS respectively) would be used for training and the one left out is used for testing. For example, C+D+MS implies that the method is trained on C, D and MS and tested on M. Only the training partitions of the source datasets are used for training. Four evaluations are performed, with each dataset taking turns as the target test set. Following the protocol set in M$^3$L and DEX, for CUHK03 the classic split is used for training, while the new split is used for evaluation.
Another popular way to conduct this leave-one-out evaluation is to merge the train, query and gallery partitions of the source datasets to form a larger training set, keeping the test set the same as before. We evaluate our \ourMethodAbrv{} on the former benchmark. However, in our experiments we compare \ourMethodAbrv{} against other methods that used the latter evaluation with more training data. 

\noindent\textbf{Single Domain Supervised Person ReID Benchmarks} Our single domain benchmarks are more straightforward, with each of the four Person ReID datasets used individually. In all experiments, we use mean average precision (mAP) and/or Rank-1 to quantify performance.
\subsection{Implementation Details}
In all experiments, we use ResNet-50 as a base model and apply cross-entropy, triplet and center losses to all subheads with $\alpha_1=1$, $\alpha_2=1$ and $\alpha_3=0.0005$. The learning rate starts at $1.75\text{e-}6$ and linearly warms up to $1.75\text{e-}4$ over the first 10 epochs. It is later reduced by a factor of 0.1 each in epochs 30 and 55. Applying Fully-Combinatorial-IN of depth $\delta=3$, we deploy $2^3=8$ subheads of 3 bottlenecks, each expressing a unique IN-pattern (see Figure~\ref{fig-IN-recipes} for an example of \ourMethodAbrv{} with $\delta=2$). We use a batch size of 32, apply random erasing~\citep{Zhong2017RandomAugmentation} with p=0.1, and color-jitter augmentation. We train for 150 epochs. For dimension reduction, we evaluate both random projections (RP) and PCA, leading to two models \ourMethodAbrv{}-RP and \ourMethodAbrv{}-PCA.

\subsection{Results for Domain Generalization Person ReID}
Table~\ref{tab:dg-reid-modern} compares recent DG-ReID methods with ours on the leave-one-out benchmark evaluations. \ourMethodAbrv{} outperforms previous SOTA by 2.4\% in Rank-1 and 3.2\% in mAP for C+M+MS $\rightarrow$ D. For D+M+MS $\rightarrow$ C, we see a jump of 2.1\% in Rank-1 and 5.0\% in mAP. Our method compares favorably against recent SOTA methods that merge the train and test sets of the source domains, showing that \ourMethodAbrv{} learns efficiently from data. It is also noteworthy that random projections perform very well with only a slight drop in performance.
\subsection{Results on Single-Domain Supervised Person ReID benchmarks}
Table~\ref{tab:reid-single-domain} compares \ourMethodAbrv{} against SOTA cross-domain and single-domain ReID methods on the simple single-domain supervised Person ReID benchmarks. Methods developed for DG-ReID perform poorly when adapted back to the single domain task because they may have been over-specialized to work only on the DG-ReID setting. The simple design and strong theoretical underpinnings of \ourMethodAbrv{} allow it to avoid this trap,  achieving or even outperforming SOTA performance in all single-domain Person ReID benchmarks. 

\subsection{Ablation Studies}
\begin{table}[h]
\caption{Ablation over major components of \ourMethodAbrv{}. Starting with a baseline, we extend it to a self-ensemble of 8 branches of depth $\delta=3$ each and then apply Fully-Combinatorial-IN to the branches.}
\center
\vspace{-1em}
\resizebox{0.9\columnwidth}{!}{
\begin{tabular}{l|cc|cc} \hline
\multicolumn{1}{c|}{\multirow{2}{*}{\textbf{Config}}} & \multicolumn{2}{c|}{\textbf{C+D+M $\rightarrow$ MS}} & \multicolumn{2}{c}{\textbf{MSMT17}} \\
\multicolumn{1}{c|}{}                                 & \textbf{mAP}              & \textbf{R1}              & \textbf{mAP}      & \textbf{R1}     \\ \hline
\textbf{Baseline}                               & 17.4                      & 42.2                     & 46.7              & 74.6            \\
\textbf{+Self-Ens (8 Branch)}                         & 19.7                      & 46.5                     & 50.0              & 77.5            \\
\textbf{+FullComb-IN ($\delta=3$)}                            & \textbf{20.8}                      & \textbf{48.7}                     & \textbf{61.0}              & \textbf{83.8}     \\ \hline      
\end{tabular}
}
\label{tab:ablation-ourmethod-components}
\end{table}
\subsubsection{\textbf{Components of \ourMethodAbrv{}}} Starting with a baseline model, a self-ensemble (Self-Ens) framework is applied to create eight branches, each with a depth of $\delta=3$. Next, we manifest Fully-Combinatorial-IN (FullComb-IN) by adding or removing IN for each bottleneck. Table~\ref{tab:ablation-ourmethod-components} compares the performance as we add features of \ourMethodAbrv{} to a baseline, showing that self-ensembles alone do contribute to performance, but Fully-Combinatorial-IN reaps more performance benefits through feature diversity.

\begin{table}[h]
\caption{Ablation on \ourMethodAbrv{} depth $\delta$. For brevity, we apply random projection to $d=2048$ and report the mAP scores.}
\vspace{-1.0em}
\center
\resizebox{0.8\columnwidth}{!}{
\begin{tabular}{c|ccccc} \hline
\multirow{2}{*}{\textbf{Benchmark}} & \multicolumn{5}{c}{\textbf{\ourMethodAbrv{} Depth ($\delta$)}}           \\
                                    & 0    & 1    & 2    & 3             & 4             \\ \hline
\textbf{DukeMTMC-reID}                 & 84.1 & 85.9 & 87.1 & 87.3          & \textbf{88.0} \\
\textbf{C → M}                      & 67.6 & 71.0 & 72.6 & \textbf{74.2} & 72.6          \\
\textbf{D+MS → C}                   & 20.6 & 22.8 & 23.4 & 25.4          & \textbf{25.6} \\
\textbf{C+D+M → MS}                 & 42.2 & 44.3 & 46.9 & \textbf{47.5} & 46            \\ \hline
\textbf{Average}                    & 53.6 & 56.0 & 57.5 & \textbf{58.6} & 58.1   \\ \hline      
\end{tabular}
}
\label{tab:ablation-ourmethod-depth}
\end{table}
\subsubsection{\textbf{Varying depths of \ourMethodAbrv{}}} Table~\ref{tab:ablation-ourmethod-depth} compares the effects of Fully-Combinatorial-IN for varying depths $\delta$ of \ourMethodAbrv{}. $\delta=0$ is the performance of a baseline model with no self-ensemble nor IN pattern diversity. As we increase depth, performance improvements are observed that eventually hit diminishing returns. Since the number of branches expressed is exponential in the depth, the computation cost of train a deeper \ourMethodAbrv{} model grows accordingly, outweighing the diminishing benefits from more IN pattern diversity. We found $\delta=3$ to be ideal for \ourMethodAbrv{}.

\subsubsection{\textbf{Dimension Reduction Study.}} 
\begin{figure}[h]
\centering
\includegraphics[width=1.0\columnwidth]{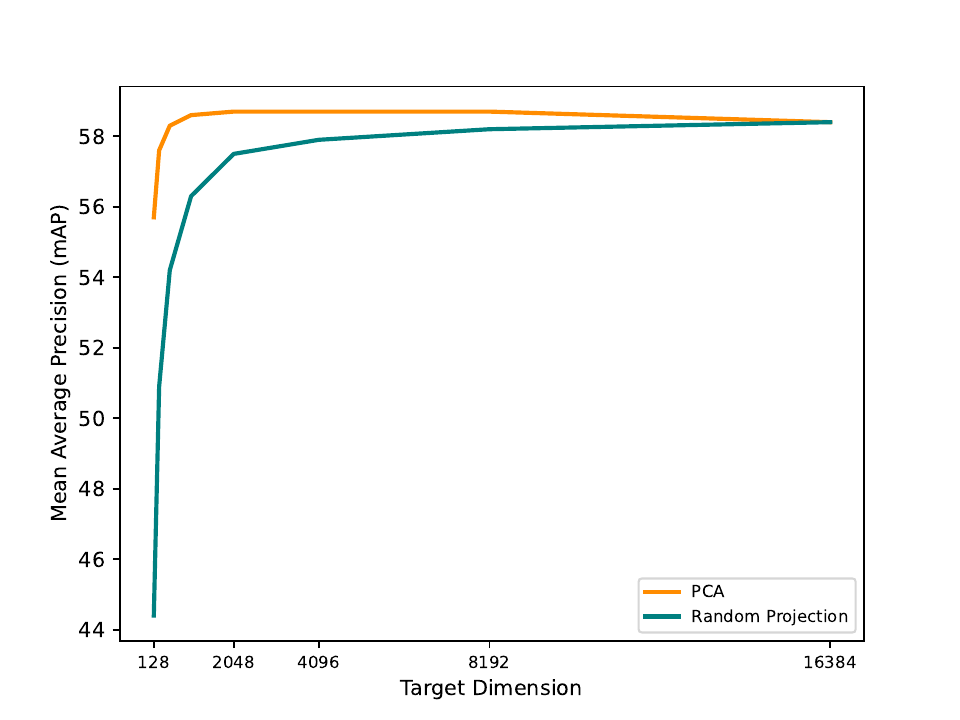} 
\vspace{-1.7em}
\caption{Comparing PCA vs. Random Projection performance on the C+D+MS $\rightarrow$ M benchmark as target dimensions are reduced. The original dimension is 16384. Down to a target dimension of around 2048, both methods are comparable.}
\vspace{-0.5em}
\label{fig-randproj-pca-chart}
\end{figure}

Figure~\ref{fig-randproj-pca-chart} compares the performance of random projections and PCA as we reduce to smaller dimensions in D $\rightarrow$ M. Using \ourMethodAbrv{} of depth $\delta=3$, the 8 subheads produce a combined feature dimension of 16384. As we reduce the dimensions of these features, their performance generally holds up well for both techniques until $d<1024$, where we see a drastic degradation of mAP for random projections. PCA, however, still continues to perform well into even lower dimensions down to $d=256$ where the mAP has dropped by \textit{only 0.8\%} from the original performance.
\vspace{-1.0em}
\begin{table}[h]
\caption{Comparing methods of dimension reduction. Values are mAP/Rank-1.}
\vspace{-1.0em}
\center
\resizebox{\columnwidth}{!}{
\begin{tabular}{c|cccc} \hline
\textbf{Benchmark} & \textbf{AutoEnc (L2)} & \textbf{AutoEnc (L1)} & \textbf{RandProj} & \textbf{PCA} \\ \hline
\textbf{Market-1501} & 85.7/94.7 & \textbf{86.0/94.9} & 84.6/94.5 & 85.5/94.8 \\
\textbf{CUHK03} & 
89.3/92.5 & \textbf{89.4/93.2} & 86.2/90.4 & 87.1/89.2 \\
\textbf{D $\rightarrow$ M} & 37.8/67.4 & 38.4/68.6 & 40.8/72.1 & \textbf{42.1/72.1} \\
\textbf{M $\rightarrow$ MS} & 10.4/28.0 & 10.1/27.3 & 12.2/32.1 & \textbf{12.8/32.7}       \\ \hline   
\end{tabular}
}
\label{tab:ablation-autoenc-vs-randproj-pca}
\end{table}
\vspace{-1em}
\subsubsection{Deep Auto-Encoder Study.} Table~\ref{tab:ablation-autoenc-vs-randproj-pca} compares the dimension reduction effectiveness between deep auto-encoders, random projections and PCA. Auto-encoder methods use a set of fully connected layers for encoding the large-dimensional feature and another set for decoding them back to the original dimension. The decoded feature is compared against the original using a reconstruction loss, which in our experiments is either a $L_1$ or $L_2$ distance loss. Auto-encoders work well only in the single-domain Person ReID tasks, but worked poorly across domains, indicating that they overfit to the source domain during training and cannot produce omni-domain generalizable encodings at test time.


\section{Conclusion}
In this work we propose the \ourMethod{} (\ourMethodAbrv{}), a method that outperforms the SOTA in both single-domain supervised Person ReID and DG-ReID benchmarks. In designing \ourMethodAbrv{} we incorporated two novel approaches. Firstly, we investigated self-ensembles in new ways to enable them for direct application in Person ReID inference. Secondly, we incorporated random projections, a technique new to Person ReID that comes from the database and text search disciplines, to efficiently and effectively reduce the large features produced by ensembles. We hope this study lays the groundwork for the development of new methods that achieve Omni-Domain Generalization in Person ReID. 

\begin{acks}
This work was supported by the Defence Science and Technology Agency (DSTA) Postgraduate Scholarship, of which Eugene P.W. Ang is a recipient. It was carried out at the Rapid-Rich Object Search (ROSE) Lab at the Nanyang Technological University, Singapore.
\end{acks}

\bibliographystyle{ACM-Reference-Format}










\end{sloppypar}
\end{document}